\pdfoutput=1

\documentclass[11pt]{article}

\usepackage{acl}

\usepackage{times}
\usepackage{latexsym}

\usepackage[T1]{fontenc}

\usepackage[utf8]{inputenc}

\usepackage{microtype}

\usepackage{graphicx}
\usepackage{subfigure}
\usepackage{booktabs} 

\usepackage{xcolor}

\definecolor{darkgreen}{rgb}{0,0.5,0}
\definecolor{orange}{rgb}{1,0.5,0}
\definecolor{teal}{rgb}{0,0.5,0.5}
\definecolor{darkpurple}{rgb}{0.5, 0, 0.5}

\newcommand {\colone}[1]{{\color{darkgreen} #1\normalfont}}
\newcommand {\coltwo}[1]{{\color{darkpurple}#1\normalfont}}

\usepackage{amsmath}
\usepackage{amssymb}
\usepackage{mathtools}
\usepackage{amsthm}

\usepackage{algorithm}
\usepackage{algorithmic}

\usepackage{hyperref}

\title{{\sc \textbf{GenAudit}}: Fixing Factual Errors in Language Model Outputs \\with Evidence}

\author{Kundan Krishna$^\clubsuit$ \quad Sanjana Ramprasad$^\diamondsuit$ \quad  Prakhar Gupta$^\clubsuit$  \\  \textbf{Byron C. Wallace}$^\diamondsuit$ 
\quad \textbf{Zachary C. Lipton}$^{\clubsuit}$ \quad \textbf{Jeffrey P. Bigham}$^{\clubsuit}$\\
$^\clubsuit$ Carnegie Mellon University \\
$^\diamondsuit$Northeastern University \\
\texttt{\small \{kundank,prakharg,zlipton,jbigham\}@andrew.cmu.edu} \\
\texttt{\small \{ramprasad.sa,b.wallace\}@northeastern.edu}}

\begin{document}
\maketitle
\begin{abstract}
LLMs can generate factually incorrect statements 
even when provided access to reference documents.
Such errors can be 
dangerous in 
high-stakes applications (e.g., document-grounded QA for healthcare or finance). 
We present {\sc GenAudit} --- a tool intended to assist fact-checking LLM responses for document-grounded tasks.
{\sc GenAudit} suggests edits 
to the LLM response
by revising or removing claims that are not supported by the reference document, 
and also presents evidence from the reference for facts that do appear to have support. 
We train 
models to 
execute these tasks, and design an interactive interface to present 
suggested edits and evidence to users. 
Comprehensive evaluation by human raters 
shows that {\sc GenAudit} can detect errors in 8 different LLM outputs when summarizing documents from diverse domains. 
User studies demonstrate that using {\sc GenAudit} can substantially improve the performance of humans at finding errors in LLM-generated summaries.
We release our tool ({\sc GenAudit}) and fact-checking model for public use.\footnote{\url{https://genaudit.org}}

\end{abstract}

\section{Introduction}
\label{sec:intro}

\begin{figure*}[ht!]
    \centering
    \includegraphics[width=\textwidth]{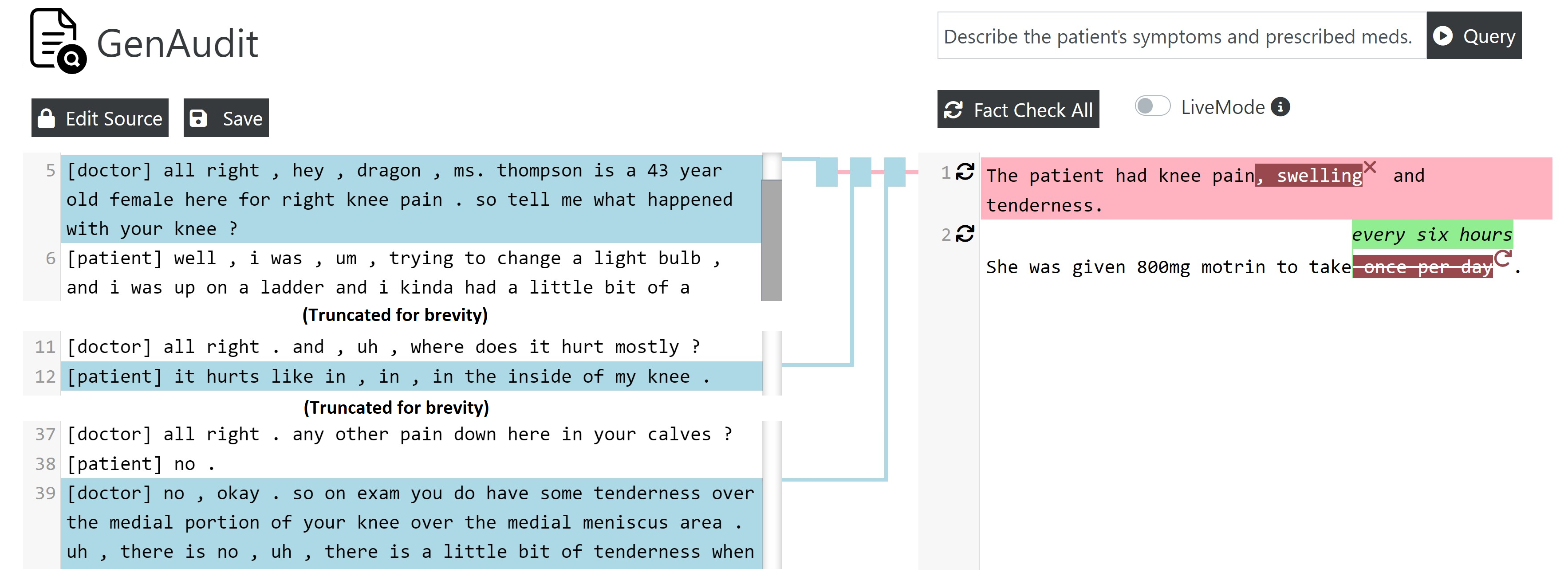}
    \caption{An illustration of {\sc GenAudit}'s user interface and sample predictions. Reference document (a clinical transcript) is on the left and the generated text to be fact-checked is on the right (generated by querying any LLM, but manually entered here for ease of illustration). Spans in the text which are not supported or are contradicted by the reference are highlighted in red, with suggested replacements in green. As the user moves to any line in the generated text, evidence found for all facts in it are highlighted using blue links. Evidence and error predictions shown here are made by a fine-tuned Flan-UL2 model backend.}
    \label{fig:intro_fig1}
\end{figure*}

LLMs 
can produce factually incorrect or unsubstantiated statements~\citep{li2023halueval,min-etal-2023-factscore}, even when they are explicitly provided relevant context such as documents~\citep{adams2023meta, sadat2023delucionqa}.
Incidentally, such \emph{document-grounded} generation is often involved in high-stakes usage scenarios where factual correctness is 
(especially) paramount. 
For example, a doctor using an LLM to
summarize a patient's medical history~\citep{adams2023meta,kanwal2022attention} might make an incorrect decision if the generated summary contains errors. 
Manually verifying LLM outputs in such settings is therefore prudent, but also time-consuming and so undercuts the motivation for using language technologies in the first place. 
This motivates the need for a system that can \emph{assist} users in efficiently verifying LLM output.

To this end, we introduce {\sc GenAudit}, a tool for fact-checking LLM responses in document-grounded tasks such as summarization and question answering.
Given a document and an LLM-generated output conditioned on the same, {\sc GenAudit} (i) locates factual errors in the output text and proposes edits to fix them, and (ii) displays evidence to support facts in the (potentially edited) text.
The system consists of two components: an interactive interface which presents evidence and edit suggestions for the user to act upon, and a bespoke backend model (fine-tuned LLM) capable of producing edits and identifying evidence.
The interface 
allows the user 
to make edits to the LLM-generated text,
and then observe updated predictions from the fact-checking model.
Notably,
in addition to supporting the task of fact-checking itself, the interface can also be used as a tool 
to evaluate and compare different backend fact-checking models, collecting data on human 
edits to 
fine-tune better models, and carry out counterfactual testing of fact-checking models by editing source documents.

We designed and evaluated different models to generate the fact-checking predictions for the tool, including fine-tuned and few-shot prompted LLMs.
We treat this as a sequence-to-sequence task: Given an  input document and a claim sentence, the model is required to simultaneously generate the sentence ids in the document which provide evidence, and a \emph{revised} version of the claim which fixes any factual errors. 
We used data from the USB benchmark~\citep{krishna-etal-2023-usb} to train and evaluate models on the fact-checking tasks. 
We found that fine-tuned open-source LLMs perform better than few-shot prompted ChatGPT and GPT-4 models, at least when evaluated on an in-domain held-out test set.

Ideally, a fact-checking tool would
support verifying text  
produced by any LLM, based on reference documents from any domain.
We evaluated {\sc GenAudit} 
using 8 different models to summarize documents from 3 different domains.
Human annotators were asked to accept or reject edits suggested by the tool, fix errors that were not caught by it, and also to provide feedback on the usefulness of suggested evidence.
On average, {\sc GenAudit} highlighted $\sim$$40\%$ of erroneous words in summaries with a precision of $\sim$$95\%$.\footnote{For reference, $\sim$$4\%$ of words in summaries are erroneous, on average.} 
In terms of extracting useful evidence,
{\sc GenAudit} achieved $\sim$$91\%$ recall and  $\sim$$95\%$ precision.

Human evaluations also showed that {\sc GenAudit} can be used to verify summarization outputs in different domains, including clinical conversations, news articles and social media posts. 
This is despite 
the fact-checking model being trained only on Wikipedia data.
{\sc GenAudit} 
successfully identified errors in outputs from 8 different LLMs including Mistral-7B~\citep{jiang2023mistral}, LLama2-70B~\citep{touvron2023llama}, Gemini-pro~\citep{team2023gemini} and GPT-4~\citep{achiam2023gpt} in human evaluation. 
Observed precision ranged between $79-100\%$ while, recall ranged from $23-57\%$ for different generation models.

We conducted a user study to measure the impact of assistance from {\sc GenAudit} on the performance of humans trying to find errors in LLM-generated summaries.
Each user fact-checked a number of summaries of news articles against the corresponding references, and for half of them they were provided assistance by {\sc GenAudit}.
We found that the percentage of erroneous words detected by the users was $62.2\%$ when they did not use {\sc GenAudit}, and it increased to $79.4\%$ when they used {\sc GenAudit}. 
This demonstrates the ability of the tool to boost performance of humans at fact-checking.

Users fact-checking LLM outputs would not want to miss any factual errors, and may prefer that that most potential errors are highlighted for review, even if some of them are false alarms.
In terms of modeling, this translates to giving a higher importance to getting better recall than precision at error-detection.
We therefore introduce a
decoding algorithm for fact-checking models which generate revised/fixed versions of claims, which can increase the recall of error detection while minimizing drop in precision. 
This approach entails intervening at time-steps where the output probabilities fall below a threshold $\tau$ to select alternate decoding paths.
Varying $\tau$ allows us to suggest more or fewer edits according to user's preference, 
effectively trading recall against precision.
This approach produces a 
better precision-recall frontier compared to baselines.

\begin{table*}[t!]
\centering
\resizebox{0.8\textwidth}{!}{
    \begin{tabular}{p{\textwidth}}
\toprule
\textbf{Input:}\\
You are provided a document and its summary. The summary may potentially contain factual errors. The last sentence of the summary is marked as a claim. Find all sentences in the document providing evidence for the claim, and then revise the claim to remove or replace unsupported facts.\\
DOCUMENT: 
SENT0 Micheal Ward 
SENT1 Early life. 
SENT2 Micheal Ward was born in Spanish Town, Jamaica on 18 November 1997. 
SENT3 His mother was 18 years old when he was born. 
SENT4 He has three sisters. 
...
SENT17 Ward's breakout year came in 2019, when he starred as Jamie in Netflix's revival and third series of "Top Boy".
\colone{SENT18 He also appeared in a leading role in the film "Blue Story" in the same year.} 
SENT19 The film received critical acclaim, and Ward won the BAFTA Rising Star Award for his performance. 
...\\
SUMMARY: Micheal Ward (born 18 November 1997) is a Jamaican-British actor and former model. \\
CLAIM: \coltwo{His films include "Blue Story" (2018) and "The Old Guard" (2020).} \\
\midrule
\textbf{Output:} \\
EVIDENCE: \colone{SENT18} \\
REVISION: \coltwo{His films include "Blue Story".} \\
\bottomrule
    \end{tabular}
}
    \caption{Sample datapoint with input-target formatting from the USB dataset}
    \label{tab:usb_example}
\end{table*}

Our contributions are as follows:
\begin{itemize}
    \item We present {\sc GenAudit}, a tool to assist fact-checking LLM outputs in document-grounded tasks. The tool identifies and fixes errors, and highlights evidence for claims.
    \item We create fine-tuned LLMs which serve as backend models for fact-checking and evaluate them comprehensively on the task of fact-checking summaries generated by 8 LLMs for documents from 3 domains.
    \item We demonstrate via a user study, that using assistance from {\sc GenAudit} improves performance of humans at finding errors in LLM-generated summaries.
    \item We present and evaluate a custom decoding algorithm that allows one to improve error detection recall while incurring a smaller drop in precision than baselines. 
\end{itemize}

\section{Background}
\label{sec:background}

We fine-tune LLMs to perform evidence extraction and claim editing, and use them as backend for {\sc GenAudit}.
In this section we provide an overview of the training dataset and models we use to power the underlying evidence extraction and factual error correction tasks.

\subsection{The USB dataset}

The USB dataset~\citep{krishna-etal-2023-usb} is composed of Wikipedia articles, their summaries and (human) annotations on them.
The summaries have two versions: (i) An initial version which
may have content that is unsupported by the article or contradicted by it, and (ii) An  
edited version which annotators have created by making minimal edits to the initial version to remove errors.
Additionally, each sentence in the 
edited summary is linked to a minimal set of article sentences that provide sufficient evidence for all facts that it contains. 

We format the dataset in a sequence-to-sequence format to use it for fine-tuning LLMs (Table~\ref{tab:usb_example}). The input to the model starts with the task instruction. It is followed by the reference document where each sentence is prefixed by a sentence ID (e.g. SENT1, SENT2...). It is then followed by the summary sentences upto the sentence to be fact-checked (called the \emph{claim}).
The sentences preceding the claim are included so that relevant context from it (e.g. coreferences) can be used for better understanding of the claim.
Finally, the claim is appended to the input.
The target output consists of the two parts. The first part contains a list of sentence ids from the document which provide evidence for the claim, and the second part consists of a revised version of the claim which removes its unsupported information and replaces incorrect facts.

We use a custom split of the USB dataset for training and evaluating our model. 
We shuffle and divide the entire dataset into train, validation and test splits of size $94\%$, $3\%$ and $3\%$ of the full dataset.
This differs from the original USB splits in two ways.
First, the training split is much larger at $94\%$ instead of the original $40\%$. 
Second, the training split consists of articles from all 6 domains in the benchmark, whereas originally 2 of the domains where reserved as challenging OOD examples to occur in only the test set.
The motivation for both of these changes is that we want to create a tool which generalizes to other diverse data beyond simply Wikipedia articles, and hence we train on as much and as diverse data as possible.

\subsection{Reducing memory requirement for training}

We aim to fine-tune large models for the fact-checking task, since they are more likely to generalize better to unseen domains due to internal knowledge.
Additionally, we need to feed in the entire reference document to the model, which can be thousands of tokens long.
Both these factors increase the memory requirement for training the models, and to address this challenge, we use low-rank adapters with 4-bit quantization~\citep{dettmers2023qlora}.
Low rank adapters \citep{hu2021lora} reduce the memory requirement during training by reducing the number of trainable parameters, thus reducing gradient computation.
To reduce the sequence length in cases where the reference document (Wikipedia article) is too long, we iteratively drop sections in it which are not relevant to any sentence in the summary (i.e. do not provide any evidence for it). We follow this process until the input sequence length is reduced to within a maximum limit, which is kept the same regardless of what model we are fine-tuning.
We also use gradient accumulation and gradient checkpointing~\citep{chen2016training} to reduce memory footprint for training. For more details, please see the Appendix.

\begin{figure*}[t!]
    \centering
    \includegraphics[width=\linewidth]{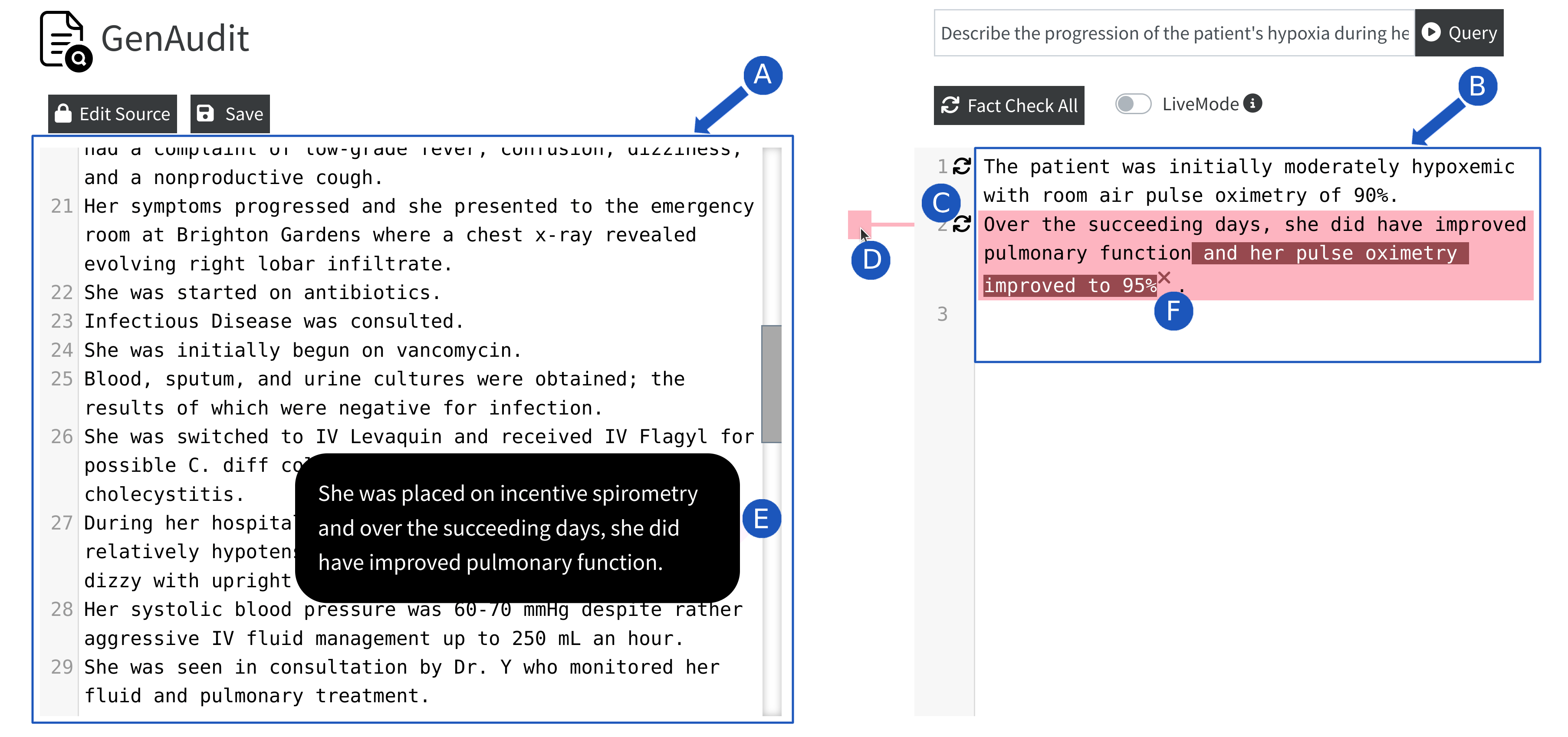}
    \caption{Overview of the Genaudit interface. The reference document (A) is displayed alongside the model generated output (B), allowing for easy comparison. Users can initiate the fact-checking process by clicking the fact-checking icon next to each output sentence (C), which highlights potential factual inconsistencies. Evidence markers (D) indicate supporting or contradicting evidence from the reference document. Hovering over an evidence marker highlights the relevant section in the reference (E), and clicking on it scrolls to the corresponding location in the document. Hallucinated spans in the generated output are marked in red (F), with the system providing suggestions for real-time correction}
    \label{fig:system_overview}
\end{figure*}
\section{System Overview}

The {\sc GenAudit} interface presented in Figure 
\ref{fig:system_overview}, allows users to interact with both the generated output and the corresponding reference document. This interface enables users to perform the following functions aimed at enhancing the factual accuracy of generated text.

\begin{itemize}
    \item Review generated output: The generated text (labeled B) is displayed adjacent to the reference document (labeled A). This side-by-side layout enables an intuitive comparison between the generated text and the document it is derived from, serving as a first step in the fact-checking process.
    
    \item Fact-Check outputs: A key feature of the interface is the ability to fact-check generated outputs with a single action. Users can fact-check any sentence in the generated text by clicking the corresponding icon next to it  (labeled C), which triggers an automated verification process that cross-references information in the generated output against the reference document. Users can also click on the ``Fact Check All'' button to trigger fact-checking of all generated sentences.
    
    \item Highlight Evidence: The system displays evidence markers (labeled D) next to each generated sentence. These markers link to the corresponding sentences of the reference document that provide supporting or contradicting evidence. When users hover over these evidence blobs, the relevant sentence of the reference document is highlighted (labeled E) to make the connection clear. Clicking on the evidence marker automatically scrolls the reference text to bring the supporting content into view, helping users quickly locate relevant evidence.
    
    \item Editable Text: In cases where factual inaccuracies are detected, { \sc GenAudit} offers automatic edit suggestions to correct the hallucinated spans. Users can interact with these proposed changes by reviewing the marked hallucinated spans (highlighted in red and labeled F). The system allows users to accept or reject the suggested edits, enabling real-time refinement of the generated text. 
    
\end{itemize}

\section{Experiments}
\label{sec:exp}

\begin{table*}
\centering
\resizebox{0.7\textwidth}{!}{
\begin{tabular}{l|ccc|ccc}
\toprule
\textbf{Model} & \multicolumn{3}{c}{\textbf{Error Identification}} & \multicolumn{3}{c}{\textbf{Evidence}}\\
{} &  Recall &  Precision &  F1 &  Recall &  Precision &  F1 \\
\midrule
\multicolumn{7}{c}{\textbf{Finetuned decoder-only LLMs}} \\
\midrule
Falcon-7B &              69.03 &                 61.54 &          65.07 &            59.85 &               54.23 &        56.90 \\
Llama2-7B &              74.85 &                 39.19 &          51.44 &            68.03 &               68.47 &        68.25 \\
Mistral-7B &              80.53 &                 73.34 &          76.77 &            72.25 &               86.66 &        78.80 \\
\midrule
\multicolumn{7}{c}{\textbf{Fine-tuned encoder-decoder LLMs}} \\
\midrule
Flan-T5-XL                     &              73.01 &                 87.07 &          79.42 &            78.90 &               85.69 &        82.16 \\
Flan-T5-XXL                    &              \textbf{80.38} &                 84.50 &          \textbf{82.39} &            \textbf{81.46} &               85.96 &        \textbf{83.65} \\
Flan-UL2       &              76.47 &                 \textbf{87.44} &          81.59 &            80.56 &               \textbf{86.42} &        83.39 \\
\midrule
\multicolumn{7}{c}{\textbf{Few-shot prompted proprietary LLMs}} \\
\midrule
GPT-3.5-turbo (8shot) & 38.79 & 48.57 & 43.13 & 51.79 & 45.15 & 48.24 \\
GPT-4 (4shot) & 37.98 & 63.89 & 47.64 & 74.42 & 38.52 & 50.76 \\
\bottomrule
\end{tabular}
}
\caption{Performance of different models on the test split of the USB dataset for the tasks of (i) identifying erroneous words, and (ii) highlighting relevant evidence}
\label{tab:usb_results}
\end{table*}

\begin{table*}[ht!]
\centering
\resizebox{1.0\textwidth}{!}{
\begin{tabular}{l|cccc|c|cccc}
\toprule
{} & \multicolumn{4}{c}{\textbf{Error Identification}} &  \multicolumn{1}{c}{\textbf{Replacements}} & \multicolumn{4}{c}{\textbf{Evidence Extraction}}\\
{} & BaseRate &  Recall &  Precision &  F1 &  Accepted\%  & Recall &  Precision &  F1 & Sufficient\% \\
\midrule
Aggregate           &                 3.97 &              40.37 &                 95.04 &          56.66 &                          78.18 &            90.83 &               95.22 &        92.97 &                     85.98 \\
\midrule
\multicolumn{10}{c}{\textbf{Summary generation models}} \\
\midrule
Llama2-7B   &                 4.29 &              30.21 &                 89.29 &          45.15 &                          66.67 &            90.71 &               96.12 &        93.33 &                     86.65 \\
Mistral-7B &                 1.99 &              23.83 &                 92.00 &          37.86 &                          40.00 &            91.43 &               95.80 &        93.57 &                     87.59 \\
Falcon-7B  &                21.84 &              47.95 &                 97.46 &          64.28 &                          86.36 &            84.65 &               87.08 &        85.85 &                     76.77 \\
Llama2-70B  &                 3.29 &              43.38 &                 95.93 &          59.75 &                         100.00 &            91.98 &               92.09 &        92.04 &                     86.10 \\
Flan-UL2   &                 9.68 &              34.04 &                 96.04 &          50.26 &                          80.00 &            90.18 &               93.96 &        92.03 &                     84.16 \\
Gemini-pro &                 1.80 &              27.75 &                 78.69 &          41.03 &                          66.67 &            91.27 &               96.98 &        94.04 &                     86.68 \\
GPT-3.5-turbo   &                 1.10 &              29.06 &                 89.47 &          43.87 &                          75.00 &            92.32 &               97.23 &        94.71 &                     88.09 \\
GPT4      &                 2.53 &              56.77 &                100.00 &          72.42 &                          87.50 &            90.39 &               97.07 &        93.61 &                     85.49 \\
\midrule
\multicolumn{10}{c}{\textbf{Datasets}}\\
\midrule
XSum          &                 4.54 &              55.00 &                 98.69 &          70.64 &                          77.78 &            94.81 &               97.92 &        96.34 &                     92.83 \\
ACIBench      &                 2.73 &              44.04 &                 90.65 &          59.28 &                          88.89 &            87.57 &               93.30 &        90.34 &                     80.84 \\
Reddit        &                 4.88 &              22.52 &                 92.41 &          36.22 &                          60.00 &            91.93 &               95.50 &        93.68 &                     86.55 \\
\bottomrule
\end{tabular}
}
\caption{Results from human evaluation of {\sc GenAudit} predictions (using fine-tuned Flan-UL2 model). Aggregate results show performance on all evaluated document-summary pairs, whereas other rows show performance (a) on summaries generated using individual LLMs, or (b) on documents from specific datasets. BaseRate represents the percentage of tokens in summaries that are hallucinations. Sufficient\% represents the percentage of summary sentences for which the evidence shown by the model was sufficient for deciding factuality.}
\label{tab:humaneval}
\end{table*}

We use the USB dataset for training/prompting and evaluating 8 different models for two factchecking tasks. 
For the evidence extraction task, we report precision, recall and F1 score as in a binary classification task where given a claim and reference document, each sentence in the reference is classified as relevant evidence or not. 
To evaluate the model's ability to remove errors, we compare the words removed/replaced by the model vs those removed in the ground truth revision.
Given the original claim and a revision, we tokenize each into words and compute the diff between them (using Python's showdiff library). 
The words in the claim that are removed/replaced in the revision are tagged as incorrect and the remaining words are tagged as correct. 
We use the ground truth revision in the dataset to compute the ground truth tags, the model-generated revision to compute the predicted tags, and compute the corresponding precision, recall and F1 scores.
Notably, it is difficult to compare the replacement text proposed by the model with ground truth replacements automatically, since the underlying text span being replaced must match exactly to make an aligned comparison.
This requires human evaluation which we discuss in the next section.

We fine-tune and evaluate 6 different models. 
These include decoder-only models from the Falcon~\citep{almazrouei2023falcon}, Llama2~\citep{touvron2023llama} and Mistral~\citep{jiang2023mistral} series, and encoder-decoder Flan-T5 models ~\citep{chung2022scaling, tay2022ul2}.
Finally, we also use OpenAI's GPT-3.5-turbo and GPT-4 models for the task via few-shot prompting, using 8 and 4 exemplars respectively.

We find that there is large variation in performance of decoder-only LLMs (Table~\ref{tab:usb_results}). 
Llama2  outperforms Falcon in evidence extraction, but underperforms it in the claim editing task, due to its low precision.
Mistral outperforms both Falcon and Llama2 models by a large margin.
There is relatively less variation in performance of the three encoder-decoder models. 
Flan-T5-XXL and Flan-UL2 models perform the best, with the former providing better recall and the latter providing better precision on both tasks.
Few-shot prompted GPT models perform worse than all fine-tuned models on both error identification and evidence extraction.

\section{Human Evaluation}
\label{sec:humaneval}

In the previous section we saw that models fine-tuned on the USB dataset perform well when evaluated on its test split. 
However, this does not imply that they would also perform well when deployed in diverse out-of-domain scenarios. 
Two 
types of domain shift can occur here.
The first is a change in the domain of reference documents 
used for fact-checking. 
USB consisted of Wikipedia articles only, but we would ideally want 
a finetuned model to work with other document types such as news articles or meeting transcripts.
Another sort of domain shift 
is the specific model generating the content to be fact-checked.
USB consists only of claims written by humans, but we would want models to detect and fix errors in content generated by a 
arbitrary LLMs.

We run experiments to evaluate the performance of {\sc GenAudit} when fact-checking summaries generated by different models for documents 
sampled from different domains. 
We include a diverse set of open-source models, including three decoder-only 7B parameter LLMs (Mistral~\citep{jiang2023mistral}, Llama2~\citep{touvron2023llama}, Falcon~\citep{almazrouei2023falcon}), one large 70B parameter model (Llama2), and one encoder-decoder model (Flan-UL2~\citep{tay2022ul2}).
As proprietary API-based models, we use GPT-3.5-turbo, GPT-4, and Gemini-pro models for summary generation.
We then use the Flan-UL2 model finetuned 
on USB to fact-check the generated summaries.

We select 
documents for summary generation from the following three datasets.

\noindent \textbf{XSum}~\citep{narayan2018don} \quad 
A summarization dataset consisting of BBC news articles 
covering diverse topics and events.

\noindent \textbf{ACI-Bench}~\citep{yim2023aci} \quad 
A dataset for summarization of patient visits to the doctor 
comprising transcripts of doctor-patient encounters. 

\noindent \textbf{Reddit-TIFU}~\citep{kim2019abstractive} \quad 
A dataset consisting of posts from the online discussion forum Reddit in which users narrate 
personal day-to-day experiences.

We randomly select 30 documents from each of the three datasets for which to generate summaries.  
For the Reddit-TIFU dataset, we 
manually filtered out examples
containing profanity or sexually explicit content. 
While generating 
summaries with open-source models, we decode using top-$p$ nucleus sampling~\citep{holtzman2019curious} from the output token distribution with a top-$p$ value of $0.9$ and a temperature of $1.0$.

We 
hired annotators via Upwork,\footnote{\url{https://www.upwork.com/}} and instructed them to
evaluate all edits suggested by the fact-checker, accept 
those that fix 
legitimate factual errors, and mark incorrect suggestions.
Annotators were also instructed to find any missing errors in
summaries which were not highlighted by the system, and to fix them by making minimal edits.

To provide feedback on the highlighted evidence, annotators provided binary (relevant/not relevant) feedback for each suggested evidence sentence in the summary.
They were also instructed to consider if the highlighted evidence for each summary sentence is sufficient or not; 
of not, then they should mark additional source sentences which contain the missing evidence. 
Additionally, we asked annotators to flag incomprehensible summaries, which were then excluded from the analysis.
For example, instead of a summary, sometimes Falcon-7B model outputs a continuation of instructions, such as \emph{``when
creating a summary, use the information given and 
avoid superfluous details.''}
The Appendix includes additional evaluation details.

Results are shown in Table~\ref{tab:humaneval}. 
We use the metrics described in Section~\ref{sec:exp} for rating 
suggested errors and evidence generated by {\sc GenAudit}.
On an aggregate level---across all domains and summary generation models---{\sc GenAudit} identifies
erroneous words with high precision ($95.04\%$) and moderate recall 
($40.37\%$). 
Note that achieving high recall is challenging here given the low prevalence of erroneous words ($3.97\%$).
With respect to evidence extraction, we 
observe 
high precision and recall ($95.22\%$ and $90.83\%$, respectively), suggesting
that most evidence sentences highlighted by the model are useful for fact-checking the given claim, and only few evidence sentences are missed (not highlighted) by the model. 
We have provided some samples of model-predicted errors in the Appendix, with the full collection available online\footnote{\url{https://aftersunrise.pythonanywhere.com} (anonymized)} .

The rate of errors in outputs varies considerably across models. 
Summaries from 
GPT-3.5-turbo have the lower error rate at $1.10\%$, while Falcon-7B has the highest error rate of $21.9\%$. 
The highest recall and precision for error detection is observed for the latter model. 
The precision of error detection remains around or above $90\%$ for all models except Gemini-pro.
Recall varies widely ($\sim$$23-57\%$) across different models.
The lowest recall is for 
Mistral-7B 
($23.83\%$); for context, its error rate is  $1.99\%$.
For evidence extraction, the performance with most models is quite similar with both precision and recall, falling between $\sim$$85-97\%$. 
The lowest F1 score is $85.85\%$ for Falcon-7B, 
and highest is $94.71\%$ for GPT-3.5-turbo.

Among the
datasets considered, {\sc GenAudit}'s performance at error identification was
best for XSum (news articles), followed by ACIBench (clinical conversations), and finally Reddit/TiFU (social media posts).
While the precision stays above $90\%$ on all datasets, the recall ranges from $22.52\%$ on the Reddit dataset to $55.00\%$ on 
XSum. 
On the evidence extraction task, {\sc GenAudit} achieves F1 scores of $90\%+$ on all three datasets.

While the previously discussed metrics measure success at identifying parts of the text which are incorrect, we also measure the quality of model-generated replacements when they are suggested.
The percent of model-suggested replacements accepted was $\sim$$78\%$, on average (Table~\ref{tab:humaneval}), suggesting the quality of generated replacement strings.
The percent of generated summary sentences for which the highlighted evidence was sufficient for verification was $\sim$$86\%$, indicating that generated evidence highlights may make fact-checking more efficient.

Using the annotations collected above, 
we evaluate 
additional models on the error detection task.
Few-shot prompted GPT-4 achieves better recall than Flan-UL2, while other fine-tuned models achieve lower recall (Table~\ref{tab:comparemodels_humaneval}).
All models achieve
lower precision than 
Flan-UL2. 
Edits suggested by GPT-4 add about $8.8\%$ more words to them on average, while other fine-tuned models add a negligible amount, reflecting the tendency of GPT-4 to make substantial changes to  text.
Finally, we also evaluate the FAVA model trained by \citet{mishra2024fine} for factual error correction. 
We see that the model achieves the lowest recall compared to all the models evaluated, with a slightly higher precision than GPT-4.

\begin{table}[]
    \centering
\resizebox{0.5\textwidth}{!}{
    \begin{tabular}{lccccc}
    \toprule
       \textbf{Model} & \textbf{Recall} & \textbf{Precision} & \textbf{\%Del} & \textbf{\%Add}\\
    \midrule
       Flan-UL2  &  40.37 & \textbf{95.04}  & 1.69 & 0.18 \\
       Flan-T5-XL  & 25.75 & 74.23 & 1.38 & 0.12 \\
       Mistral-7B & 35.08 & 45.24 & 3.08 & 0.16\\
       \midrule
       GPT-4 (4-shot) & \textbf{40.68} & 28.50 & 5.67 & 8.80\\
       FAVA~\citep{mishra2024fine} & 14.18 & 31.34 & 1.80 & 0.43 \\
    \bottomrule
    \end{tabular}
}
    \caption{Performance of models fine-tuned by us on the USB dataset, few-shot prompted GPT-4, and the FAVA model~\citep{mishra2024fine} at identifying erroneous words in model-generated summaries, along with the percentage of summary words deleted and added by their edits.}
    \label{tab:comparemodels_humaneval}
\end{table}

\section{Impact on human performance at fact-checking}
\label{sec:efficiency_eval}

In the previous section, we asked humans to evaluate the correctness of the factual errors detected by {\sc GenAudit}, as well as the relevance of evidence highlighted by it. While those experiments show how accurate {\sc GenAudit} is, it does not reveal how helpful it would be to a human carrying out the fact-checking exercise by themselves.
In this section, we measure how the fact-checking performance of a human user would change if they were given access to {\sc GenAudit}.

We recruited human subjects to carry out fact-checking of summaries, and measured their performance with and without access to {\sc GenAudit} predictions. In each task, users were given 2 article-summary pairs and were asked to find as many errors as they can in them. For one summary, they were shown the highlighted evidence and suggested edits from {\sc GenAudit}, and for the other summary they weren't shown the outputs of {\sc GenAudit}. The rest of the interface was exactly the same. 
The time available to the human subject for fact-checking a summary can heavily influence their performance, and performance can perhaps increase indefinitely as more time is provided.
To control for this aspect, we asked the subjects to carry out the fact-checking exercise under time constraints. We gave them 3 minutes to fact-check each summary.

The subjects were recruited using a qualification task where each candidate was given two article-summary pairs to fact-check --- one with help from {\sc GenAudit} and one without. The annotation exercise was attempted by $23$ candidates. We chose the top $4$ performing candidates, where their performance was measured by the F1-score of error detection.

To create the data for this experiment, we took articles from the XSUM dataset and generated summaries using GPT-3.5-turbo model. Since the rate of errors in summaries generated are quite low (ref. Table~\ref{tab:humaneval}), most of the summaries won't have any errors. This can be problematic, because we would like to have a relatively large number of errors to get a reliable estimate of subjects' performance at error detection. Hence, we chose to inject errors in each summary synthetically by prompting GPT-3.5-Turbo again, following the process outlined in \citet{laban2023summedits} (more details are provided in the Appendix). This provided us with a larger pool of errors, as well as ground truth labels marking the erroneous spans in the summaries post-error-addition.

We got $96$ document-summary pairs fact-checked by the subjects who cleared our qualification round, out of which $48$ document-summary pairs had {\sc GenAudit}-produced annotations to help in the fact-checking.
We found that for the summaries where {\sc GenAudit}-produced annotations were not given, the subjects achieved an aggregate recall of $62.2\%$ at finding the errors. In contrast, for the summaries where {\sc GenAudit}-produced annotations were given, the recall was $79.4\%$.
Hence, the availability of {\sc GenAudit} enabled subjects to find a higher percentage of factual errors in the summaries. 
The difference between the two groups of summaries where {\sc GenAudit} was used vs not,  measured in terms of the the recall achieved by the subjects, was statisfically significant according to Student's t-test ($\text{p-value} < 0.02$).

\section{Improving Recall of Error Detection}
\label{sec:tuning}

Users fact-checking LLM outputs using {\sc GenAudit} may give more importance to a higher recall than precision to be confident that most errors are highlighted for review, even at the cost of false positives. 
While it is always possible to increase recall by indiscriminately flagging additional text spans as errors, a naive strategy would lead to a large drop in precision.
We propose a 
decoding algorithm for the fact-checking model which uses the output token probabilities to achieve a better precision-recall trade-off.

Our proposed approach (Algorithm~\ref{algo:threshedit}) for increasing error detection recall 
relies on observing the probabilities of tokens generated as the revision by the fact-checker, and intervening at timesteps with low model confidence. 
Given a document $D$, claim $C$, and an initially generated revision $R=r_1r_2..r_m$, we find the first position $t$ where the probability 
assigned to token $r_t$ falls below a 
threshold $\tau$.
At that timestep we then generate the token with the highest probability \emph{excluding} $r_t$.
We generate the remaining tokens (from $t+1$) as usual via greedy decoding to compute an alternate revision $R'$.
Given $R$ and $R'$, assume the span $r_kr_{k+1}..r_{k+w}$ was replaced by $x_1..x_{q}$.
We make the replacement in $R$ yielding $r_1...r_{k-1}x_1..x_{q}r_{k+w+1}...r_{m}$.
We then repeat the process of finding low probability tokens and making edits for the remaining tokens.
After each iteration of the while loop, the value of $(|R|-t)$ decreases by at least 1, which guarantees termination of the program.

\begin{algorithm*}[tb]
   \caption{Thresholded Edit}
   \label{algo:threshedit}
\begin{algorithmic}
   \STATE {\bfseries Input:} document $D$, claim $C$, predicted evidence $E$, predicted revision $R$, model $\mathcal{M}$, threshold $\tau$
   \STATE $Q=(D,C,E)$
   \STATE $t=1$
   \WHILE{$t \leq |R|$}
   \STATE{{\small//~{\color{blue} run this loop until the timestep counter reaches end of string R}\normalsize}}
   \STATE $p_1p_2...p_{|V|}$ = ${\text{\bfseries NextTokProb}}_{\mathcal{M}}(r_1..r_{t-1} \mid Q)$
   \COMMENT{run M to get probability of tokens that can be generated at timestep t}
   \IF{$p_{r_{t}} \leq \tau$}
   \STATE{{\small//~{\color{blue} if the probability of the token currently at timestep t in R is lower than tau, we replace it}\normalsize}}
   \STATE $r' = \arg\max_{k}( p_k \mid k\neq r_{t})$
   \COMMENT{replace it with the token with the next highest probability (call it r')}
   \STATE $\text{prefix} = r_1r_2..r_{t-1}r'$
   \STATE $\text{compl}$ = ${\text{\bfseries 
   Generate}}_{\mathcal{M}}(\text{prefix} \mid Q)$
   \COMMENT{use M to generate the full completion of the revision if we use r' at timestep t}
   \STATE $R' = \text{prefix} + \text{compl}$
   \COMMENT{we call the new revision R'}
   \STATE $N_{del}, N_{add}, \text{repl}$ = $ \text{{\bfseries DiffAtPos}}_{t}(R,R')$
   \COMMENT{extract span-level change between R and R' at timestep t (deletions/replacements)}
   \STATE $R = \text{prefix} + \text{repl} + r_{(t+N_{del})}...r_{|R|}$
   \COMMENT{commit that change at t in R and discard any other changes observed later}
   \STATE $t = t + N_{add}$  
   \ENDIF
   \STATE $t = t + 1$
   \COMMENT{advance the pointer and repeat the loop to find more spans to edit in R if they exist}
   \ENDWHILE
   \STATE {\bfseries Output:} updated revision $R$
\end{algorithmic}
\end{algorithm*}

\begin{figure}
    \centering
\includegraphics[width=0.5\textwidth]{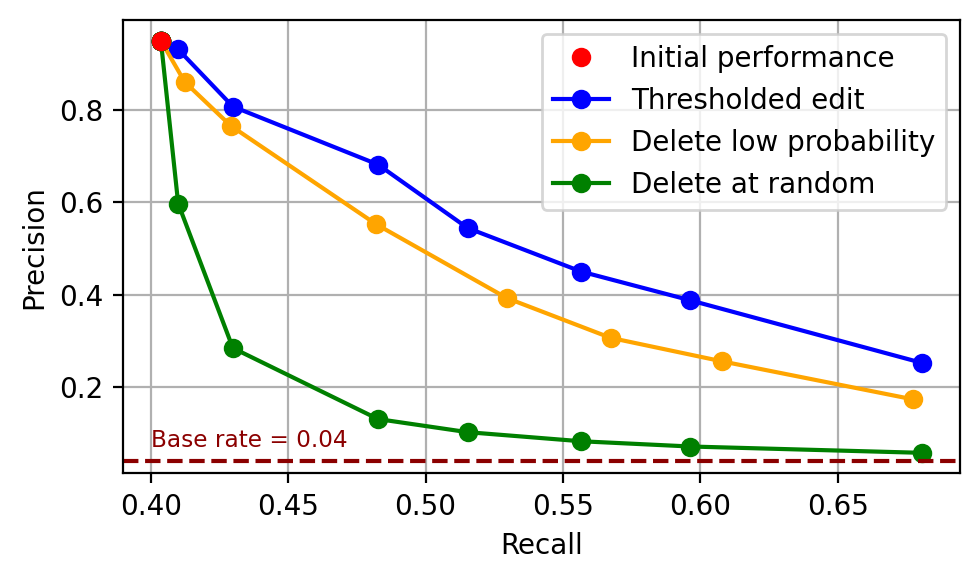}
    \caption{Variation in precision and recall of error identification by a fine-tuned Flan-UL2 model when using thresholded editing (Algorithm~\ref{algo:threshedit}), versus editing out additional tokens either at random or by selecting the ones with low probability.}
    \label{fig:precrecall}
\end{figure}

Increasing the value of $\tau$ in Algorithm~\ref{algo:threshedit} would lead to more edits being made to the claim, and vice-versa. 
We run the Flan-UL2 fact-checking model with different values of $\tau$ ranging from $0.0$ to $0.99$
and plot the resulting recall and precision at detecting errors annotated in the human evaluation experiment in Section~\ref{sec:humaneval}.
We find that using Algorithm~\ref{algo:threshedit}, we are able to increase the recall from about $40\%$ to $60\%$, with a drop in precision from $95\%$ to $40\%$.
Although there is a drop in precision, the drop is much lower than what one would get by using a simple randomized baseline.
The baseline strategy we compare against is to boost recall by randomly selecting additional words (beyond the ones already predicted as erroneous by the model) and mark them as erroneous too.
We compute the number of words that need to be selected to boost expected recall to a certain level, and the resulting drop in expected precision that it entails (see Appendix for derivation).
The thresholding approach maintains a much higher precision with increasing recall compared to the baseline strategy, where the precision already falls to around $28\%$ when the recall increases to merely $43\%$ (Figure~\ref{fig:precrecall}).

We also compare using the custom decoding strategy in Algorithm~\ref{algo:threshedit} 
with simply flagging additional tokens as non-factual if their probability of generation (by the fact-checking model) falls below a 
variable threshold. 
We see that this strategy performs worse than using Algorithm~\ref{algo:threshedit} (Figure~\ref{fig:precrecall}).
This suggests that post-hoc usage of token probabilities from the fact-checking model does not isolate the non-factual spans as well as active intervention during the decoding process as done in Algorithm~\ref{algo:threshedit}.

\section{Related Work}

With the rapid rise in capabilities of large language models~\citep{achiam2023gpt,touvron2023llama,jiang2023mistral}, they have been widely used for assisting humans in content generation workflows such as writing summaries~\citep{jung2023interactive}, or scientific articles~\citep{singh2024figura11y}.
However, it has been known for a long time that such language models produce factually inaccurate statements, popularly known as hallucinations~\citep{arthur2016incorporating,see2017get,cao2018faithful}.
This leads to a lack of trust in the AI-generated assistance, and multiple works have noted this limitation~\citep{qian2024take, swaroop2024accuracy}.
To ensure factual accuracy of text, some prior works have designed intelligent user interfaces. 
\citet{rony2020claimviz} introduced a tool which helps journalists fact-check claims made in debates by retrieving relevant evidence via web-search. 
In contrast, our work also predicts whether claims are true or false in addition to highlighting relevant evidence from reference document.
Concurrent work from \citet{fu2024data} proposed a tool where claims in text are verified against structured data tables. 
In contrast, our system works with free-form text as reference, and does not need access to data tables whose creation can be complex and may capture only specific categories of information.

Multiple works in the natural language processing literature have introduced approaches to predict whether a generated summary contains any factual error, either using trained models~\citep{kryscinski2020evaluating,goyal2021annotating}, question-answering based approaches~\citep{fabbri2022qafacteval},
or prompting LLMs~\citep{laban2023summedits}.
While these works predicted factual correctness with respect to a given source document, recent works have implemented fact-checking against a large corpus (Wikipedia) by combining evidence retrieval and factuality prediction~\citep{kamoi-etal-2023-wice, min-etal-2023-factscore}.
Our effort goes beyond binary prediction of factual correctness, by also localizing the errors in the claims and fixing them via minimal editing. \citet{mishra2024fine} also attempts to fix factual errors via editing, and we compared the performance of their released model with ours in Section~\ref{sec:humaneval}.

\citet{liu-etal-2023-improving} and \citet{krishna-etal-2023-usb} introduced the DeFacto and USB datasets respectively with human annotations to train and evaluate models for revising incorrect claims, and extracting evidence from a reference document.
While both datasets can potentially be used to train {\sc GenAudit} backend models, we used the USB dataset because (1) it contains comprehensive evidence labels for \emph{all} facts in the claim, and (2) it contains multi-sentence summaries,
which are more common in practice.
We extend these lines of work by contributing an interactive tool for fact-checking, and a comprehensive evaluation of the models trained on such data at fixing errors in modern LLM outputs with evidence.

\section{Conclusion}

We introduced {\sc GenAudit}, a tool to assist users in fact-checking LLM generated outputs against inputs by presenting supporting evidence and highlighting (and fixing) errors.
We trained models for fact-checking tasks which rival few-shot prompting of SOTA LLMs, and designed a web-interface for users to interact with.
We evaluated {\sc GenAudit} for fact-checking summaries generated by 8 LLMs for documents in 3 domains.
We demonstrated, via a user study, that using {\sc GenAudit} improves performance of humans at fact-checking summaries.
Finally, we proposed a decoding algorithm for our fact-checking model to improve the recall of error identification while minimizing the cost in precision.

\section*{Acknowledgements}
This work was supported in part by the National Science Foundation (NSF), grant RI: 2211954 and 2211955.

\bibliography{main}

\begin{thebibliography}{40}
\expandafter\ifx\csname natexlab\endcsname\relax\def\natexlab#1{#1}\fi

\bibitem[{Achiam et~al.(2023)Achiam, Adler, Agarwal, Ahmad, Akkaya, Aleman, Almeida, Altenschmidt, Altman, Anadkat et~al.}]{achiam2023gpt}
Josh Achiam, Steven Adler, Sandhini Agarwal, Lama Ahmad, Ilge Akkaya, Florencia~Leoni Aleman, Diogo Almeida, Janko Altenschmidt, Sam Altman, Shyamal Anadkat, et~al. 2023.
\newblock Gpt-4 technical report.
\newblock \emph{arXiv preprint arXiv:2303.08774}.

\bibitem[{Adams et~al.(2023)Adams, Zuckerg, and Elhadad}]{adams2023meta}
Griffin Adams, Jason Zuckerg, and No{\'e}mie Elhadad. 2023.
\newblock A meta-evaluation of faithfulness metrics for long-form hospital-course summarization.
\newblock In \emph{Machine Learning for Healthcare Conference}, pages 2--30. PMLR.

\bibitem[{Almazrouei et~al.(2023)Almazrouei, Alobeidli, Alshamsi, Cappelli, Cojocaru, Debbah, Goffinet, Hesslow, Launay, Malartic et~al.}]{almazrouei2023falcon}
Ebtesam Almazrouei, Hamza Alobeidli, Abdulaziz Alshamsi, Alessandro Cappelli, Ruxandra Cojocaru, M{\'e}rouane Debbah, {\'E}tienne Goffinet, Daniel Hesslow, Julien Launay, Quentin Malartic, et~al. 2023.
\newblock The falcon series of open language models.
\newblock \emph{arXiv preprint arXiv:2311.16867}.

\bibitem[{Arthur et~al.(2016)Arthur, Neubig, and Nakamura}]{arthur2016incorporating}
Philip Arthur, Graham Neubig, and Satoshi Nakamura. 2016.
\newblock Incorporating discrete translation lexicons into neural machine translation.
\newblock In \emph{Proceedings of the 2016 Conference on Empirical Methods in Natural Language Processing}. Association for Computational Linguistics.

\bibitem[{Cao et~al.(2018)Cao, Wei, Li, and Li}]{cao2018faithful}
Ziqiang Cao, Furu Wei, Wenjie Li, and Sujian Li. 2018.
\newblock Faithful to the original: Fact aware neural abstractive summarization.
\newblock In \emph{Proceedings of the AAAI Conference on Artificial Intelligence}, volume~32.

\bibitem[{Chen et~al.(2016)Chen, Xu, Zhang, and Guestrin}]{chen2016training}
Tianqi Chen, Bing Xu, Chiyuan Zhang, and Carlos Guestrin. 2016.
\newblock Training deep nets with sublinear memory cost.
\newblock \emph{arXiv preprint arXiv:1604.06174}.

\bibitem[{Chung et~al.(2022)Chung, Hou, Longpre, Zoph, Tay, Fedus, Li, Wang, Dehghani, Brahma et~al.}]{chung2022scaling}
Hyung~Won Chung, Le~Hou, Shayne Longpre, Barret Zoph, Yi~Tay, William Fedus, Yunxuan Li, Xuezhi Wang, Mostafa Dehghani, Siddhartha Brahma, et~al. 2022.
\newblock Scaling instruction-finetuned language models.
\newblock \emph{arXiv preprint arXiv:2210.11416}.

\bibitem[{Dettmers et~al.(2023)Dettmers, Pagnoni, Holtzman, and Zettlemoyer}]{dettmers2023qlora}
Tim Dettmers, Artidoro Pagnoni, Ari Holtzman, and Luke Zettlemoyer. 2023.
\newblock Qlora: Efficient finetuning of quantized llms.
\newblock \emph{arXiv preprint arXiv:2305.14314}.

\bibitem[{Fabbri et~al.(2022)Fabbri, Wu, Liu, and Xiong}]{fabbri2022qafacteval}
Alexander~Richard Fabbri, Chien-Sheng Wu, Wenhao Liu, and Caiming Xiong. 2022.
\newblock Qafacteval: Improved qa-based factual consistency evaluation for summarization.
\newblock In \emph{Proceedings of the 2022 Conference of the North American Chapter of the Association for Computational Linguistics: Human Language Technologies}, pages 2587--2601.

\bibitem[{Fu et~al.(2024)Fu, Guo, Hoffswell, Bursztyn, Rossi, and Stasko}]{fu2024data}
Yu~Fu, Shunan Guo, Jane Hoffswell, Victor~S Bursztyn, Ryan Rossi, and John Stasko. 2024.
\newblock " the data says otherwise"-towards automated fact-checking and communication of data claims.
\newblock \emph{arXiv preprint arXiv:2409.10713}.

\bibitem[{Goyal and Durrett(2021)}]{goyal2021annotating}
Tanya Goyal and Greg Durrett. 2021.
\newblock Annotating and modeling fine-grained factuality in summarization.
\newblock In \emph{Proceedings of the 2021 Conference of the North American Chapter of the Association for Computational Linguistics: Human Language Technologies}, pages 1449--1462.

\bibitem[{Holtzman et~al.(2019)Holtzman, Buys, Du, Forbes, and Choi}]{holtzman2019curious}
Ari Holtzman, Jan Buys, Li~Du, Maxwell Forbes, and Yejin Choi. 2019.
\newblock The curious case of neural text degeneration.
\newblock In \emph{International Conference on Learning Representations}.

\bibitem[{Hu et~al.(2021)Hu, Wallis, Allen-Zhu, Li, Wang, Wang, Chen et~al.}]{hu2021lora}
Edward~J Hu, Phillip Wallis, Zeyuan Allen-Zhu, Yuanzhi Li, Shean Wang, Lu~Wang, Weizhu Chen, et~al. 2021.
\newblock Lora: Low-rank adaptation of large language models.
\newblock In \emph{International Conference on Learning Representations}.

\bibitem[{Jiang et~al.(2023)Jiang, Sablayrolles, Mensch, Bamford, Chaplot, Casas, Bressand, Lengyel, Lample, Saulnier et~al.}]{jiang2023mistral}
Albert~Q Jiang, Alexandre Sablayrolles, Arthur Mensch, Chris Bamford, Devendra~Singh Chaplot, Diego de~las Casas, Florian Bressand, Gianna Lengyel, Guillaume Lample, Lucile Saulnier, et~al. 2023.
\newblock Mistral 7b.
\newblock \emph{arXiv preprint arXiv:2310.06825}.

\bibitem[{Jung et~al.(2023)Jung, Seo, Jung, Chung, Ryu, and Chang}]{jung2023interactive}
Jeesu Jung, Hyein Seo, Sangkeun Jung, Riwoo Chung, Hwijung Ryu, and Du-Seong Chang. 2023.
\newblock Interactive user interface for dialogue summarization.
\newblock In \emph{Proceedings of the 28th International Conference on Intelligent User Interfaces}, pages 934--957.

\bibitem[{Kamoi et~al.(2023)Kamoi, Goyal, Diego~Rodriguez, and Durrett}]{kamoi-etal-2023-wice}
Ryo Kamoi, Tanya Goyal, Juan Diego~Rodriguez, and Greg Durrett. 2023.
\newblock \href {https://doi.org/10.18653/v1/2023.emnlp-main.470} {{W}i{CE}: Real-world entailment for claims in {W}ikipedia}.
\newblock In \emph{Proceedings of the 2023 Conference on Empirical Methods in Natural Language Processing}, pages 7561--7583, Singapore. Association for Computational Linguistics.

\bibitem[{Kanwal and Rizzo(2022)}]{kanwal2022attention}
Neel Kanwal and Giuseppe Rizzo. 2022.
\newblock Attention-based clinical note summarization.
\newblock In \emph{Proceedings of the 37th ACM/SIGAPP Symposium on Applied Computing}, pages 813--820.

\bibitem[{Kim et~al.(2019)Kim, Kim, and Kim}]{kim2019abstractive}
Byeongchang Kim, Hyunwoo Kim, and Gunhee Kim. 2019.
\newblock Abstractive summarization of reddit posts with multi-level memory networks.
\newblock In \emph{Proceedings of the 2019 Conference of the North American Chapter of the Association for Computational Linguistics: Human Language Technologies, Volume 1 (Long and Short Papers)}, pages 2519--2531.

\bibitem[{Krishna et~al.(2023)Krishna, Gupta, Ramprasad, Wallace, Bigham, and Lipton}]{krishna-etal-2023-usb}
Kundan Krishna, Prakhar Gupta, Sanjana Ramprasad, Byron Wallace, Jeffrey Bigham, and Zachary Lipton. 2023.
\newblock {USB}: A unified summarization benchmark across tasks and domains.
\newblock In \emph{Findings of the Association for Computational Linguistics: EMNLP 2023}, pages 8826--8845.

\bibitem[{Kry{\'s}ci{\'n}ski et~al.(2020)Kry{\'s}ci{\'n}ski, McCann, Xiong, and Socher}]{kryscinski2020evaluating}
Wojciech Kry{\'s}ci{\'n}ski, Bryan McCann, Caiming Xiong, and Richard Socher. 2020.
\newblock Evaluating the factual consistency of abstractive text summarization.
\newblock In \emph{Proceedings of the 2020 Conference on Empirical Methods in Natural Language Processing (EMNLP)}, pages 9332--9346.

\bibitem[{Laban et~al.(2023)Laban, Kry{\'s}ci{\'n}ski, Agarwal, Fabbri, Xiong, Joty, and Wu}]{laban2023summedits}
Philippe Laban, Wojciech Kry{\'s}ci{\'n}ski, Divyansh Agarwal, Alexander~Richard Fabbri, Caiming Xiong, Shafiq Joty, and Chien-Sheng Wu. 2023.
\newblock Summedits: Measuring llm ability at factual reasoning through the lens of summarization.
\newblock In \emph{Proceedings of the 2023 Conference on Empirical Methods in Natural Language Processing}, pages 9662--9676.

\bibitem[{Laban et~al.(2022)Laban, Schnabel, Bennett, and Hearst}]{laban2022summac}
Philippe Laban, Tobias Schnabel, Paul~N Bennett, and Marti~A Hearst. 2022.
\newblock Summac: Re-visiting nli-based models for inconsistency detection in summarization.
\newblock \emph{Transactions of the Association for Computational Linguistics}, 10:163--177.

\bibitem[{Li et~al.(2023)Li, Cheng, Zhao, Nie, and Wen}]{li2023halueval}
Junyi Li, Xiaoxue Cheng, Wayne~Xin Zhao, Jian-Yun Nie, and Ji-Rong Wen. 2023.
\newblock Halueval: A large-scale hallucination evaluation benchmark for large language models.
\newblock In \emph{Proceedings of the 2023 Conference on Empirical Methods in Natural Language Processing}, pages 6449--6464.

\bibitem[{Liu et~al.(2023)Liu, Deb, Teruel, Halfaker, Radev, and Awadallah}]{liu-etal-2023-improving}
Yixin Liu, Budhaditya Deb, Milagro Teruel, Aaron Halfaker, Dragomir Radev, and Ahmed~Hassan Awadallah. 2023.
\newblock \href {https://doi.org/10.18653/v1/2023.acl-long.844} {On improving summarization factual consistency from natural language feedback}.
\newblock In \emph{Proceedings of the 61st Annual Meeting of the Association for Computational Linguistics (Volume 1: Long Papers)}, pages 15144--15161, Toronto, Canada. Association for Computational Linguistics.

\bibitem[{Mangrulkar et~al.(2022)Mangrulkar, Gugger, Debut, Belkada, Paul, and Bossan}]{peft}
Sourab Mangrulkar, Sylvain Gugger, Lysandre Debut, Younes Belkada, Sayak Paul, and Benjamin Bossan. 2022.
\newblock Peft: State-of-the-art parameter-efficient fine-tuning methods.
\newblock \url{https://github.com/huggingface/peft}.

\bibitem[{Min et~al.(2023)Min, Krishna, Lyu, Lewis, Yih, Koh, Iyyer, Zettlemoyer, and Hajishirzi}]{min-etal-2023-factscore}
Sewon Min, Kalpesh Krishna, Xinxi Lyu, Mike Lewis, Wen-tau Yih, Pang Koh, Mohit Iyyer, Luke Zettlemoyer, and Hannaneh Hajishirzi. 2023.
\newblock \href {https://doi.org/10.18653/v1/2023.emnlp-main.741} {{FA}ct{S}core: Fine-grained atomic evaluation of factual precision in long form text generation}.
\newblock In \emph{Proceedings of the 2023 Conference on Empirical Methods in Natural Language Processing}, pages 12076--12100, Singapore. Association for Computational Linguistics.

\bibitem[{Mishra et~al.(2024)Mishra, Asai, Balachandran, Wang, Neubig, Tsvetkov, and Hajishirzi}]{mishra2024fine}
Abhika Mishra, Akari Asai, Vidhisha Balachandran, Yizhong Wang, Graham Neubig, Yulia Tsvetkov, and Hannaneh Hajishirzi. 2024.
\newblock Fine-grained hallucination detection and editing for language models.
\newblock \emph{arXiv preprint arXiv:2401.06855}.

\bibitem[{Narayan et~al.(2018)Narayan, Cohen, and Lapata}]{narayan2018don}
Shashi Narayan, Shay~B Cohen, and Mirella Lapata. 2018.
\newblock Don’t give me the details, just the summary! topic-aware convolutional neural networks for extreme summarization.
\newblock In \emph{Proceedings of the 2018 Conference on Empirical Methods in Natural Language Processing}, pages 1797--1807.

\bibitem[{Qian and Wexler(2024)}]{qian2024take}
Crystal Qian and James Wexler. 2024.
\newblock Take it, leave it, or fix it: Measuring productivity and trust in human-ai collaboration.
\newblock In \emph{Proceedings of the 29th International Conference on Intelligent User Interfaces}, pages 370--384.

\bibitem[{Rasley et~al.(2020)Rasley, Rajbhandari, Ruwase, and He}]{rasley2020deepspeed}
Jeff Rasley, Samyam Rajbhandari, Olatunji Ruwase, and Yuxiong He. 2020.
\newblock Deepspeed: System optimizations enable training deep learning models with over 100 billion parameters.
\newblock In \emph{Proceedings of the 26th ACM SIGKDD International Conference on Knowledge Discovery \& Data Mining}, pages 3505--3506.

\bibitem[{Rony et~al.(2020)Rony, Hoque, and Hassan}]{rony2020claimviz}
Md~Main~Uddin Rony, Enamul Hoque, and Naeemul Hassan. 2020.
\newblock Claimviz: Visual analytics for identifying and verifying factual claims.
\newblock In \emph{2020 IEEE Visualization Conference (VIS)}, pages 246--250. IEEE.

\bibitem[{Sadat et~al.(2023)Sadat, Zhou, Lange, Araki, Gundroo, Wang, Menon, Parvez, and Feng}]{sadat2023delucionqa}
Mobashir Sadat, Zhengyu Zhou, Lukas Lange, Jun Araki, Arsalan Gundroo, Bingqing Wang, Rakesh Menon, Md~Parvez, and Zhe Feng. 2023.
\newblock Delucionqa: Detecting hallucinations in domain-specific question answering.
\newblock In \emph{Findings of the Association for Computational Linguistics: EMNLP 2023}, pages 822--835.

\bibitem[{See et~al.(2017)See, Liu, and Manning}]{see2017get}
Abigail See, Peter~J Liu, and Christopher~D Manning. 2017.
\newblock Get to the point: Summarization with pointer-generator networks.
\newblock In \emph{Proceedings of the 55th Annual Meeting of the Association for Computational Linguistics (Volume 1: Long Papers)}, pages 1073--1083.

\bibitem[{Singh et~al.(2024)Singh, Wang, and Bragg}]{singh2024figura11y}
Nikhil Singh, Lucy~Lu Wang, and Jonathan Bragg. 2024.
\newblock Figura11y: Ai assistance for writing scientific alt text.
\newblock In \emph{Proceedings of the 29th International Conference on Intelligent User Interfaces}, pages 886--906.

\bibitem[{Swaroop et~al.(2024)Swaroop, Bu{\c{c}}inca, Gajos, and Doshi-Velez}]{swaroop2024accuracy}
Siddharth Swaroop, Zana Bu{\c{c}}inca, Krzysztof~Z Gajos, and Finale Doshi-Velez. 2024.
\newblock Accuracy-time tradeoffs in ai-assisted decision making under time pressure.
\newblock In \emph{Proceedings of the 29th International Conference on Intelligent User Interfaces}, pages 138--154.

\bibitem[{Tay et~al.(2022)Tay, Dehghani, Tran, Garcia, Wei, Wang, Chung, Bahri, Schuster, Zheng et~al.}]{tay2022ul2}
Yi~Tay, Mostafa Dehghani, Vinh~Q Tran, Xavier Garcia, Jason Wei, Xuezhi Wang, Hyung~Won Chung, Dara Bahri, Tal Schuster, Steven Zheng, et~al. 2022.
\newblock Ul2: Unifying language learning paradigms.
\newblock In \emph{The Eleventh International Conference on Learning Representations}.

\bibitem[{Team et~al.(2023)Team, Anil, Borgeaud, Wu, Alayrac, Yu, Soricut, Schalkwyk, Dai, Hauth et~al.}]{team2023gemini}
Gemini Team, Rohan Anil, Sebastian Borgeaud, Yonghui Wu, Jean-Baptiste Alayrac, Jiahui Yu, Radu Soricut, Johan Schalkwyk, Andrew~M Dai, Anja Hauth, et~al. 2023.
\newblock Gemini: a family of highly capable multimodal models.
\newblock \emph{arXiv preprint arXiv:2312.11805}.

\bibitem[{Touvron et~al.(2023)Touvron, Martin, Stone, Albert, Almahairi, Babaei, Bashlykov, Batra, Bhargava, Bhosale et~al.}]{touvron2023llama}
Hugo Touvron, Louis Martin, Kevin Stone, Peter Albert, Amjad Almahairi, Yasmine Babaei, Nikolay Bashlykov, Soumya Batra, Prajjwal Bhargava, Shruti Bhosale, et~al. 2023.
\newblock Llama 2: Open foundation and fine-tuned chat models.
\newblock \emph{arXiv preprint arXiv:2307.09288}.

\bibitem[{Wolf et~al.(2019)Wolf, Debut, Sanh, Chaumond, Delangue, Moi, Cistac, Rault, Louf, Funtowicz et~al.}]{wolf2019huggingface}
Thomas Wolf, Lysandre Debut, Victor Sanh, Julien Chaumond, Clement Delangue, Anthony Moi, Pierric Cistac, Tim Rault, R{\'e}mi Louf, Morgan Funtowicz, et~al. 2019.
\newblock Huggingface's transformers: State-of-the-art natural language processing.
\newblock \emph{arXiv preprint arXiv:1910.03771}.

\bibitem[{Yim et~al.(2023)Yim, Fu, Ben~Abacha, Snider, Lin, and Yetisgen}]{yim2023aci}
Wen-wai Yim, Yujuan Fu, Asma Ben~Abacha, Neal Snider, Thomas Lin, and Meliha Yetisgen. 2023.
\newblock Aci-bench: a novel ambient clinical intelligence dataset for benchmarking automatic visit note generation.
\newblock \emph{Scientific Data}, 10(1):586.

\end{thebibliography}
\bibliographystyle{acl_natbib}

\appendix

\section{Appendix}
\label{sec:appendix}

\section{Binary Classification of Factuality}

In the previous sections, we evaluated the performance of {\sc GenAudit} at localizing factual errors within text and suggesting edits.
However, it can also be repurposed as a binary classifier which simply predicts whether a long-form generated text is factually consistent or not with respect to given reference.
To do that, we simply declare a given passage of text as factually inconsistent with respect to reference document if {\sc GenAudit} suggests any edit to any sentence in it.

We evaluate the performance of {\sc GenAudit} on the SummEdits benchmark~\citep{laban2023summedits} which consists of document-summary pairs where the summaries potentially contain factual errors.
The source documents are taken from 10 different datasets representing a diverse group including legal documents, scientific papers, emails etc.
We tokenize the source document into individual sentences before passing it through the fact-checking model.
{\sc GenAudit} achieves a balanced accuracy score of $74.7$, outperforming many LLMs and traditional fact-checking methods, with the exception of Gemini-pro and GPT-4 (Table~\ref{tab:summedits})\footnote{Values taken from the official Github repository \url{https://github.com/salesforce/factualNLG}}.

\begin{table}[h]
    \centering
    \resizebox{0.48\textwidth}{!}{
    \begin{tabular}{lc}
    \toprule
        \textbf{Model} & \textbf{Balanced Accuracy} \\
    \midrule
        Human Performance & 90.92 \\
        GPT4 & 82.06 \\
        Gemini-pro & 75.49 \\
        {\sc \textbf{GenAudit}} & \textbf{74.75} \\
        Claudev21 & 74.36 \\
        Claudev2 & 73.58 \\
        ChatGPT & 71.18 \\
        PaLM-bison & 69.04 \\
        QAFactEval~\citep{fabbri2022qafacteval} & 65.46 \\
        Llama2-13b & 58.35 \\
        Mistral-7b & 57.78 \\
        SummaCConv~\citep{laban2022summac} & 57.14 \\
        DAE~\citep{goyal2021annotating} & 55.17 \\
        Llama2-7b & 50.36 \\
    \bottomrule
    \end{tabular}
    }
    \caption{Performance of models on the SummEdits benchmark for binary classification of factuality. Here {\sc GenAudit} uses the fine-tuned Flan-UL2 backend, whereas other LLMs are zero-shot prompted.}
    \label{tab:summedits}
\end{table}

\subsection{Details on human evaluation (for Section~\ref{sec:humaneval})}
\label{sec:human_pay}

We engaged annotators via Upwork\footnote{\url{https://www.upwork.com/}}, leveraging their expertise to evaluate model-generated summaries against source documents. Candidates were chosen through a qualifying round, focusing on their ability to identify inaccuracies in summaries.
Ultimately, two proficient proofreaders and fact-checkers were selected based on their performance on the qualifying task. Each annotator was tasked with annotating all summaries for half of the documents in each of the 3 datasets used (see Section~\ref{sec:humaneval}). Annotators received compensation at an average rate of \$25 USD per hour for their contributions.

\begin{figure*}[ht!]
    \centering
    \includegraphics[width=\textwidth]{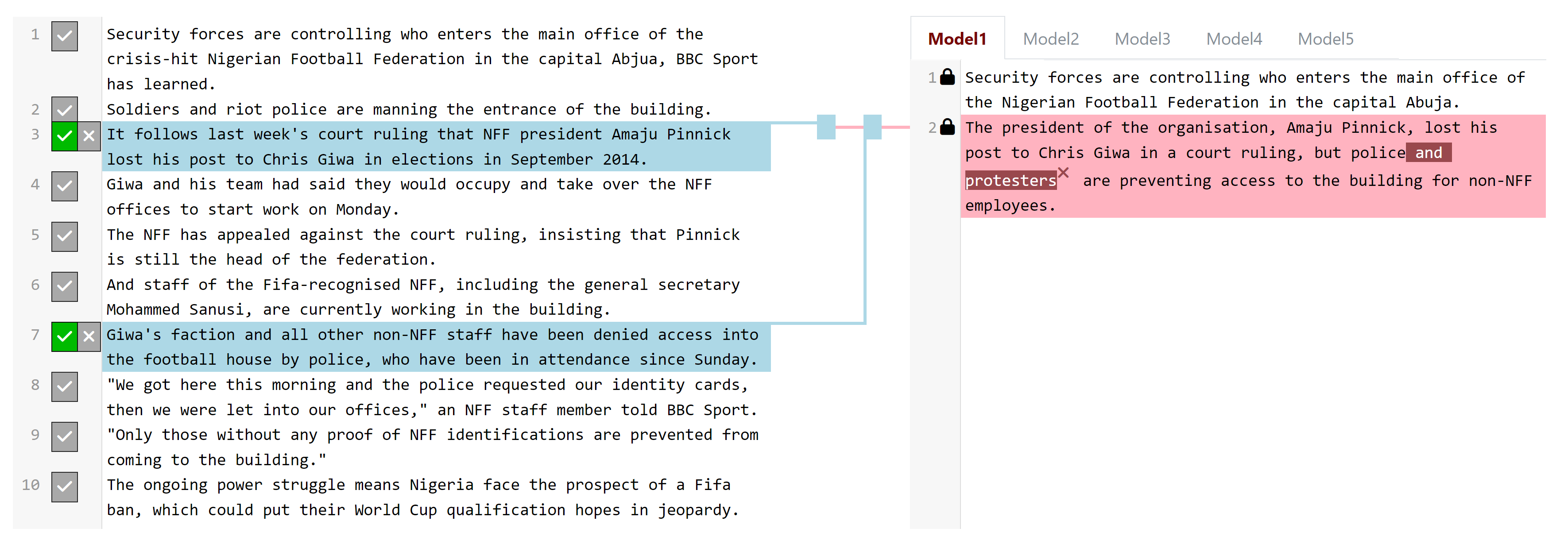}
    \caption{Interface used for collecting feedback on suggested evidence and edits from {\sc GenAudit}. Annotators can accept/reject each suggested evidence sentence, and can also mark additional sentences as evidence if needed. Suggested edits can be accepted/rejected by clicking on the button on the top-right of the highlighted span. if needed, users can also make freeform edits to fix more errors. Annotators cycle through the summaries generated by different models, whose names are anonymized and their order is shuffled.}
    \label{fig:rating_interface}
\end{figure*}

We use a slightly modified version of the {\sc GenAudit} UI to collect the annotations, with two notable changes (Figure~\ref{fig:rating_interface}). 
First, we add a buttons next to every source sentence to provide a accept/reject feedback if the source sentence is highlighted as evidence for a summary sentence. 
For source sentences which are not highlighted as evidence, we provide an accept button to mark it as additional evidence if needed.
Second, the UI enables cycling through multiple model-generated summaries for the same source document at once.
This is done to save annotators' time, since otherwise the annotators would have to read the source document again each time they get a summary for it in the sequence of annotation jobs.
The annotators were instructed to mark the following categories of generated summary sentences as \emph{invalid} which we excluded from our analysis:
(i) truncated sentences, which occur due to the maximum decode length limit being reached.
(ii) incomprehensible sentences, such as when the models generate instructions instead of summary (e.g. Falcon-7B once generated \textit{``when creating a summary, use the information given and avoid superfluous details.''})

\subsection{Annotator Instructions (for Section~\ref{sec:humaneval})}
\label{sec:annotator_instructions}

The following protocol was provided to annotators for assessing both the supporting evidence for summary sentences, and their factual consistency:

\textit{Evidence Annotation}:

\textit{a) Evaluate each summary sentence by reviewing all linked evidence from the source, marking them as either accepted (by clicking a tick) or rejected (by clicking a cross) based on their validity.}

\textit{b) Examine the summary for unsupported information with respect to suggested evidence, If any information supporting the summary is found in the source but is not already supported by the suggested evidence, mark it as "new" evidence. Note that "new" evidence pertains only to instances where the summary includes seemingly unsupported information compared to the suggested evidence.}\\

\textit{Summary Annotation}:

\textit{a) First, accept all suggestions made by the tool that are \textbf{valid}. This encompasses deletions made by the tool of unsupported information and any recommended edits or replacements.}

\textit{b) If there is incorrect or contradictory information in the summary not satisfactorily addressed by the tool then make minimal edits to the summary to align it with the source. If the edit is based on source information not already included in suggested evidence, label it as "new evidence."}

\textit{c) For cases where the summary introduces unsupported information not addressed by the tool, remove corresponding segments of the summary without altering the evidence.}

\textit{Note if a summary sentence is incomplete or  incomprehensible, mark the sentence as INVALID.}

\subsection{Inter-annotator agreement (for Section~\ref{sec:humaneval})}

For one-third of the documents in each dataset, we got their model-generated summaries annotated by both annotators to estimate inter-annotator agreement. The doubly-annotated data included $232$ summaries consisting of a total of $989$ sentences. 
We compute the agreement on error identification by comparing the words removed/replaced from the initial summary by the two annotators. 
The value of Cohen's Kappa for this is $62.7$ indicating substantial agreement.
Similarly, we compare the ratings (useful vs not) provided to each suggested evidence sentence by the two annotators, which yields a Cohen's Kappa value of $58.56$ indicating moderate agreement.

\subsection{Creating data for human fact-checking performance evaluation (for Section~\ref{sec:efficiency_eval})}

We sampled news articles randomly from the XSum dataset to use as reference text, and used GPT-3.5-Turbo to generate their summaries. We further used GPT-3.5-Turbo to modify the summaries generated in this first round by adding factual errors. We used the prompt template used by \citet{laban2023summedits} to generate the modified summaries with added errors.\footnote{\url{https://github.com/salesforce/factualNLG/blob/master/prompts/summedits/step2\_inconsistent.txt}}
The prompt template is shown in Figure~\ref{fig:erroradd_prompt}.

\begin{figure*}
    \centering
    \begin{tabular}{p{0.95\textwidth}}
{\small
    
    \begin{verbatim}

Document:
[DOCUMENT]

Consistent Summary:
[CONSISTENT_SUMMARY]

Given the document and consistent summary above, generate one slightly modified 
version of the summary such that the modifications introduce a factual 
inconsistency. For example, you can modify a number, date, or entity, and negate 
or modify a statement. Here are some rules to follow:
- The modifications should change at most 5-6 words from the original summary, 
 and keep the rest the same.
- The modifications should change  different parts of the original summary.
- Your modifications should be challenging to detect: modify minimally while 
 still introducing a factual inconsistency.
- The factual inconsistency you introduce should be subtle. For example if you 
 replace an entity, make sure you replace it with another entity from the document.
- The modified summary should start with "[FIRST_THREE_WORDS] [...]", and end 
with "[LAST_THREE_WORDS]"
    \end{verbatim}

}
    \end{tabular}
    \caption{Prompt template used for introducing factual errors in summaries used for human fact-checking performance evaluation (Section~\ref{sec:efficiency_eval}). Taken from \citet{laban2023summedits}}
    \label{fig:erroradd_prompt}
\end{figure*}

\subsection{Derivation of baseline precision-recall trade-off}

Assume a binary classification problem with a dataset of $T$ datapoints, out of which $P$ have positive labels and $N$ have negative labels. Let's assume a prediction model assigns positive labels to $P'$ datapoints, and achieves a recall of $\alpha$ and precision of $\beta$. We want to boost the recall to a target of $\alpha'$ by flipping the predictions from negative to positive for some datapoints. In this section we derive the precision-recall trade-off achieved if we select the labels to flip uniformly at random, which is the baseline used in Section~\ref{sec:tuning}.

If we flip the labels of $k$ datapoints from negative to positive, the expected number of them which would be true positives will be 
$$\Delta = \frac{P(1-\alpha)}{(T-P')}k$$

For an expected target recall of $\alpha'$, we get 
$$\alpha' = \frac{\alpha P + \Delta}{P}$$
$$\implies \alpha' = \alpha + \frac{k(1-\alpha)}{(T-P')}$$
$$\implies k = \frac{\alpha'-\alpha}{1-\alpha}(T-P')$$

The expected true positives is $\alpha'P$, which yields the new expected precision $\beta'$ to be
$$\beta' = \frac{\alpha'P}{(P'+k)}$$
$$\implies \beta' = \frac{\alpha'P}{P'+\frac{(\alpha'-\alpha)}{(1-\alpha)}(T-P')}$$

Note that $P=\gamma T$, where $\gamma$ is the base rate for positive class, and the number of predicted positive labels $P' = \frac{\alpha P}{\beta}$. Substituting these, we get our final form

$$ \beta' = \frac{\alpha' \gamma}{\frac{\alpha \gamma}{\beta}+\frac{(\alpha'-\alpha)}{(1-\alpha)}(1-\frac{\alpha \gamma}{\beta})}$$

We showed the performance of this baseline against our proposed Algorithm~\ref{algo:threshedit} in Figure~\ref{fig:precrecall}.

\subsection{Training details}
\label{sec:training_details}

Each model fine-tuned on the USB dataset, was trained for 10 epochs and the best checkpoint was selected based on validation performance at the end of each epoch.
The effective batch size was kept at 128.
The optimizer used was 8-bit AdamW with $\beta_1=0.9$ and $\beta_2=0.999$, without weight decay, and with a learning rate of $5e-5$.
The maximum input and output sequence lengths were set to $3050$ and $150$ tokens respectively.
We used 4-bit quantization with low-rank adapters~\citep{dettmers2023qlora} with parameters $r=8$ and $\alpha=32$ .

Each training run was carried out on $2\times$ Nvidia A6000 GPUs each of which has 49GB of memory. We used gradient accumulation and gradient checkpointing to reduce the peak memory requirements during training. The duration of each training run varied depending on the model used but fell in the range of 30-70 hours.
The implementation was done using the Huggingface transformers~\citep{wolf2019huggingface}, Deepspeed~\citep{rasley2020deepspeed}, Peft~\citep{peft} and Bitsandbytes\footnote{\url{https://github.com/TimDettmers/bitsandbytes}} libraries.

\begin{table*}[t]
    \centering
    \resizebox{1.0\textwidth}{!}{
    \begin{tabular}{lll}
    \toprule
        \textbf{Model} 
        & \textbf{License} & \textbf{URL}  \\
    \midrule
        Llama2-7B & llama2 & \url{https://huggingface.co/meta-llama/Llama-2-7b-chat-hf} \\
        Llama2-70B &
        llama2 & \url{https://huggingface.co/meta-llama/Llama-2-70b-chat-hf} \\
        Mistral-7B & 
        Apache-2.0 & \url{https://huggingface.co/mistralai/Mistral-7B-Instruct-v0.1} \\
        Falcon-7B & 
        Apache-2.0 & \url{https://huggingface.co/tiiuae/falcon-7b-instruct} \\
        Flan-T5-XL &
        Apache-2.0 & \url{https://huggingface.co/google/flan-t5-xl} \\
        Flan-T5-XXL &
        Apache-2.0 & \url{https://huggingface.co/google/flan-t5-xxl} \\
        Flan-UL2 & 
        Apache-2.0 & \url{https://huggingface.co/google/flan-ul2} \\
        GPT-3.5-turbo & openai & \url{https://platform.openai.com/docs/models/gpt-3-5-turbo} (version gpt-3.5-turbo-16k-0613)\\
        GPT-4 & 
        openai & \url{https://platform.openai.com/docs/models/gpt-4-and-gpt-4-turbo} (gpt-4-0613) \\
    \bottomrule
    \end{tabular}
    }
    \caption{Links to models used in Section~\ref{sec:exp} and Section~\ref{sec:humaneval}}
    \label{tab:model_urls}
\end{table*}

\begin{table*}[t]
    \centering
    \resizebox{0.72\textwidth}{!}{
    \begin{tabular}{lll}
    \toprule
        \textbf{Dataset} 
        & \textbf{License} & \textbf{URL}  \\
    \midrule
        USB & Apache-2.0 & \url{https://huggingface.co/datasets/kundank/usb} \\
        XSum & Unknown & \url{https://huggingface.co/datasets/EdinburghNLP/xsum} \\
        ACIBench & CC-BY-4.0 & \url{https://github.com/wyim/aci-bench} \\
        Reddit-TIFU & MIT & \url{https://huggingface.co/datasets/reddit_tifu} \\
        SummEdits & Apache-2.0 & \url{https://github.com/salesforce/factualNLG} \\
    \bottomrule
    \end{tabular}
    }
    \caption{Links to datasets used in this work}
    \label{tab:dataset_urls}
\end{table*}

\begin{table*}[]
    \centering
    \begin{tabular}{cp{0.8\textwidth}}
    \toprule
    \textbf{Model} & \textbf{Prompt} \\
    \midrule
    GPT-4 & You are provided a document and its summary. The summary may potentially contain facts which contradict with the document or are not supported by any evidence in the document. The last sentence of the summary is marked as a claim. Find and list sufficient sentences in the document to provide evidence for the claim. Make sure to provide evidence for all the supported facts in the claim. Then, revise the claim to remove or replace facts which are not supported by the document or are contradicted by it. Only make changes to the text of the claim when necessary. When you add new information to the claim, it must be only to fix a contradictory fact in the claim, and not for changing the style of the text. \\
    \midrule
    GPT-3.5-turbo & You are provided a document and its summary. The summary may potentially contain facts which contradict with the document or are not supported by any evidence in the document. The last sentence of the summary is marked as a claim. Find and list sufficient sentences in the document to provide evidence for the claim, and then revise the claim to remove or replace facts which are not supported by the document or are contradicted by it. When you add new information to the claim, it must be only to fix a contradictory fact in the claim, and not for changing the style of the text.\\
    \midrule
    Others & You are provided a document and its summary. The summary may potentially contain factual errors. The last sentence of the summary is marked as a claim. Find all sentences in the document providing evidence for the claim, and then revise the claim to remove or replace unsupported facts. \\
    \bottomrule
    \end{tabular}
    \caption{Prompts used for fact-checking using GPT models in zero-shot setting, and using other models with fine-tuning (as described in Section~\ref{sec:exp})}
    \label{tab:allprompts_fc}
\end{table*}

\begin{table*}[]
    \centering
    \begin{tabular}{p{0.15\textwidth}p{0.8\textwidth}}
    \toprule
    \textbf{Model} & \textbf{Prompt} \\
    \midrule
    Gemini-pro, GPT-3.5-turbo, GPT-4 & Generate a summary for the following document in brief. When creating the summary, only use information that is present in the document. Generate the summary in free-form text without using bullet points.\\
    \midrule
    Others & Generate a summary for the following document in brief. When creating the summary, only use information that is present in the document.\\
    \bottomrule
    \end{tabular}
    \caption{Prompts used for summary generation using different models for human evaluation experiment (Section~\ref{sec:humaneval})}
    \label{tab:allprompts_gensum}
\end{table*}

\begin{figure*}[h!]
    \centering
\includegraphics[width=\textwidth]{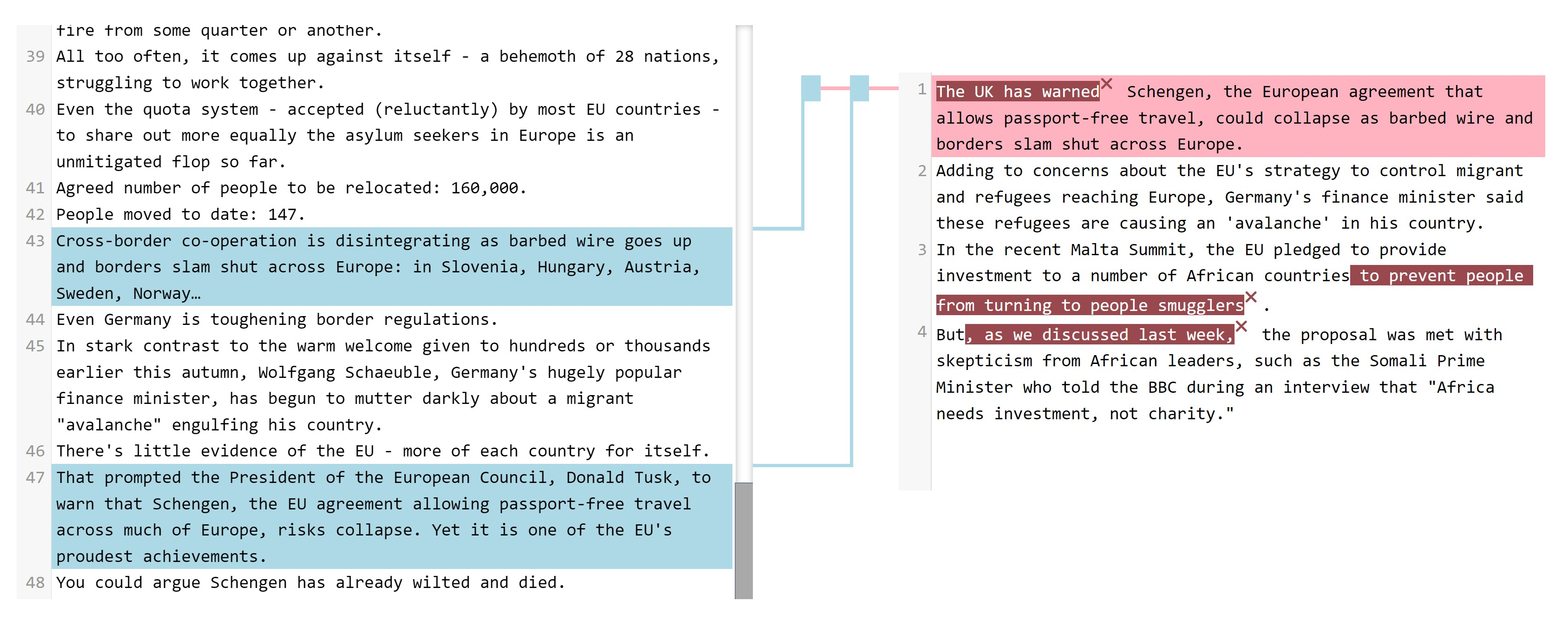}
    \caption{{\sc GenAudit} suggested edits for a GPT4-generated summary of a news article from the XSum dataset. The first sentence contains a statement attributed to the UK which was actually made by the president of the European Union}
    \label{fig:ex1}
\end{figure*}

\begin{figure*}[h!]
    \centering
\includegraphics[width=\textwidth]{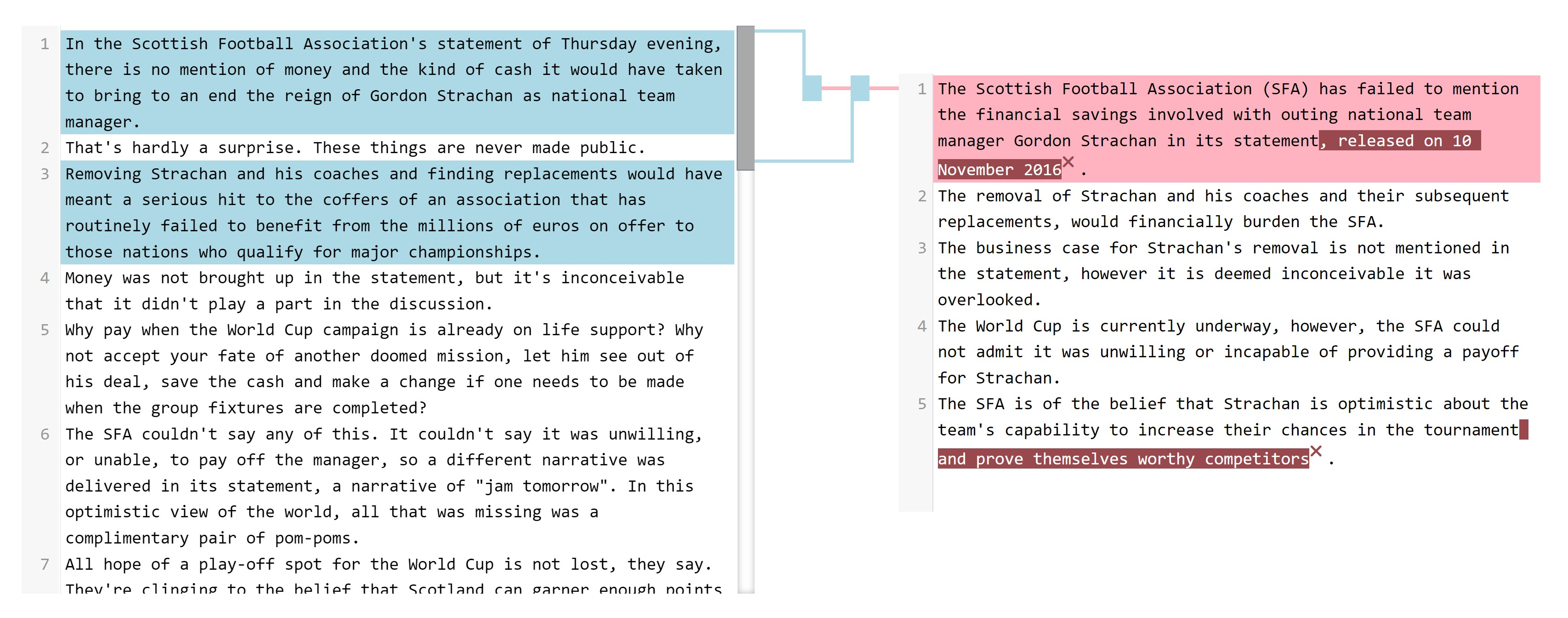}
    \caption{{\sc GenAudit} suggested edits for a GPT4-generated summary of a news article from the XSum dataset. Interestingly, The reference document does not contain the date on which the said statement was made by SFA. Interestingly though, GPT-4 almost got it right by sheer memorization. The statement was released on 17 November 2016 whereas GPT-4 mentioned 10 November 2016. {\small (Ref: \url{https://www.bbc.com/sport/football/38021627},\url{https://www.bbc.com/sport/football/38019477})}
    }
    \label{fig:ex2}
\end{figure*}

\begin{figure*}[h!]
    \centering
\includegraphics[width=\textwidth]{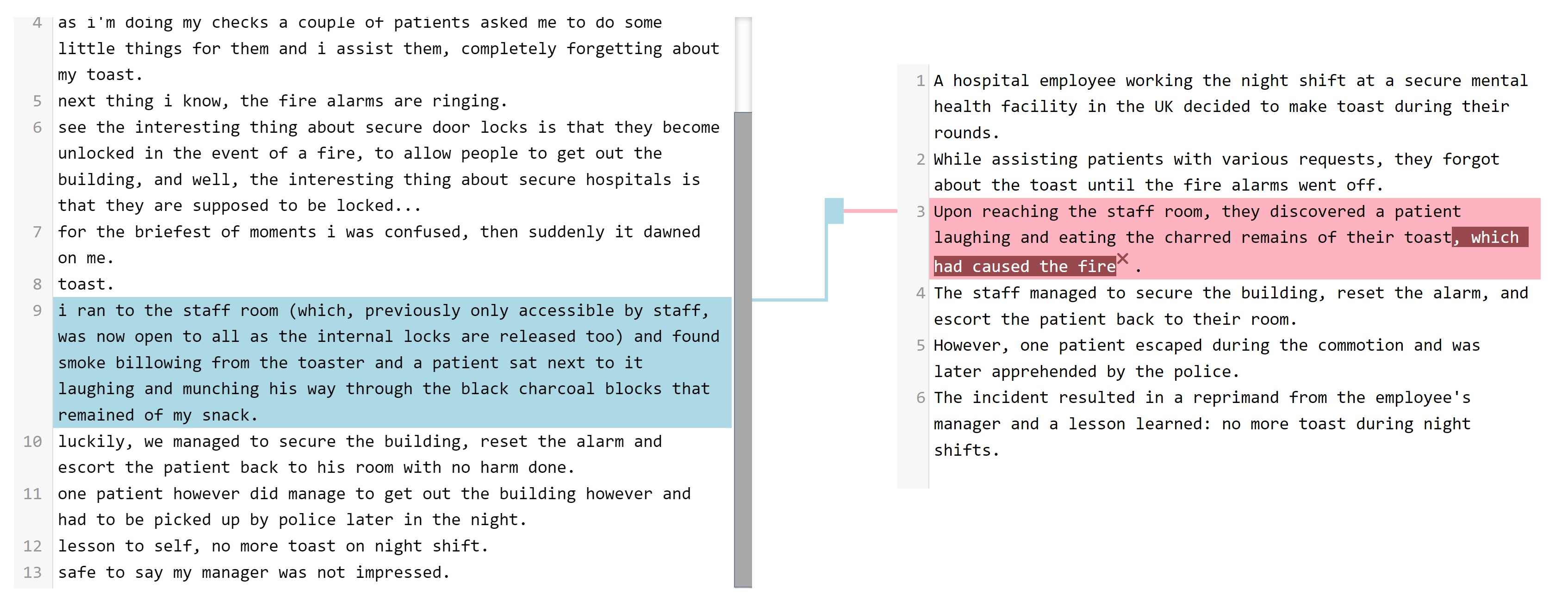}
    \caption{{\sc GenAudit} suggested edits for a Geminipro-generated summary of a Reddit post where a person describes fire alarm going off at a workplace due to smoke from burnt toast. The summary suggests that there was a fire caused which doesn't seem to be the case from the reference.
    }
    \label{fig:ex3}
\end{figure*}

\begin{figure*}[h!]
    \centering
\includegraphics[width=\textwidth]{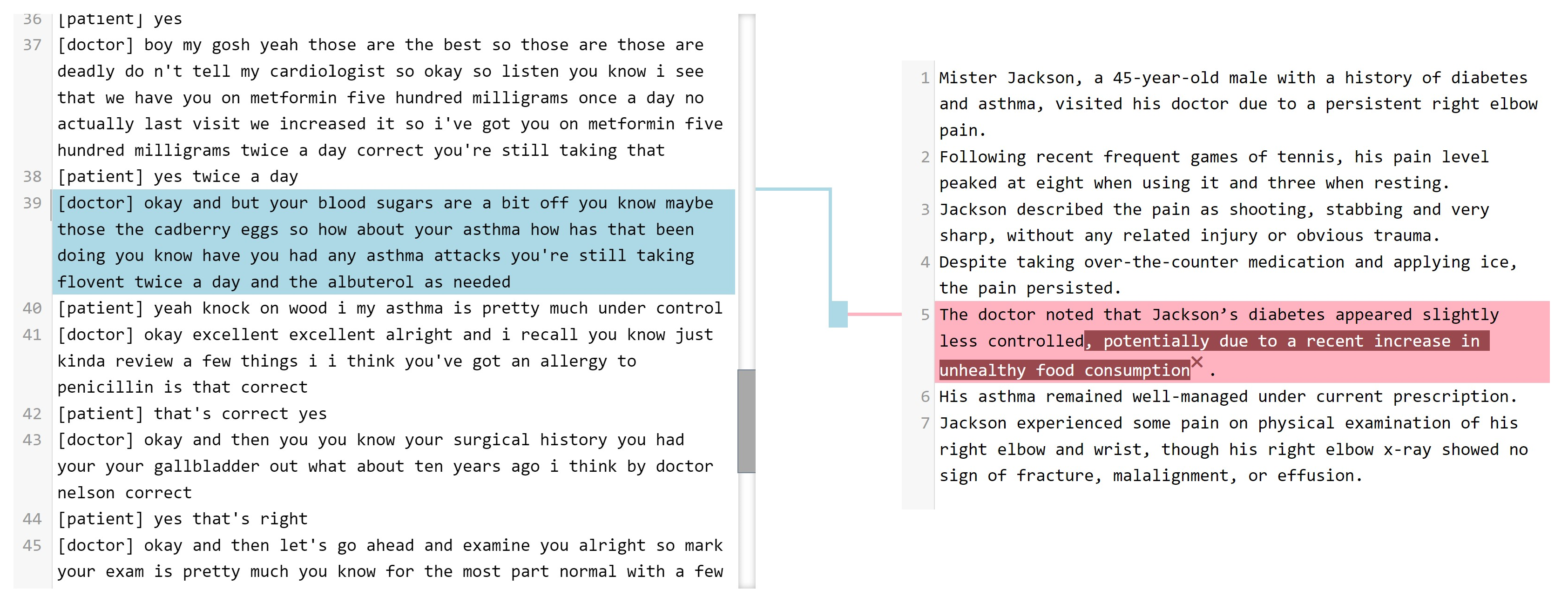}
    \caption{{\sc GenAudit} suggested edits for a GPT4-generated summary of a conversation between a doctor and patient. Here, the doctor briefly mentions that the patient's blood sugar problems may be caused by eating chocolates, but they don't suggest that such unhealthy consumption has increased recently (as the summary claims).
    }
    \label{fig:ex4}
\end{figure*}

\end{document}